\documentclass[preprint,5p,times,twocolumn, authoryear]{elsarticle}

\usepackage[colorlinks,bookmarksopen,bookmarksnumbered,citecolor=red,urlcolor=red]{hyperref}
\usepackage{booktabs}
\usepackage{array}

\journal{Environment and Planning B: Urban Analytics and City Science}

\begin{document}

\begin{frontmatter}

\title{PairWise Image Finder: An Open-source Tool for Finding Visually Aligned Street-Level Image Pairs for Urban Perception Studies}

\author[inst1]{Jussi Torkko}\corref{cor1}

\ead{jussi.torkko@helsinki.fi}
\cortext[cor1]{Corresponding author}

\affiliation[inst1]{
  organization={Digital Geography Lab, Department of Geosciences and Geography, University of Helsinki},
  addressline={PO Box 64 (Gustaf Hällströmin katu 2), 00014},
  city={Helsinki},
  country={Finland}
}

\begin{abstract}

Change detection and scene recognition techniques have been widely applied to Street View Imagery (SVI) to understand changes in scenes across the years. 
However, metadata alone is often insufficient to reliably find visually aligned image pairs.
This study introduces the PairWise image finder, a tool that integrates feature detection and matching, supported by semantic segmentation masks to quantify the visual alignment of two images of varying time periods. 
The tool outputs the share of matched key features, the matched feature distance and coverage, and the alignment of semantic masks, which enables the user to filter image pairs depending on the alignment quality and use case.
The visually aligned pairs derived from the tool can be used to accurately study explicit longitudinal change and help reduce manual effort for perception studies. 
The usability of the tool is demonstrated through a comparison of longitudinal changes, highlighting the importance of perspective when quantifying changes.
The proposed method provides a scalable and open tool for researchers and stakeholders to find high-quality image pairs for urban analysis, perception and related applications.

\end{abstract}


\end{frontmatter}

\section{Introduction}

Scene Change Detection (SCD) is the task of understanding changes between two images~\citep{choEnvironmentalChangeDetection2025,kannanZeroSCDZeroShotStreet2024}. 
It is closely related Visual Place Recognition (VPR), the task of determining whether two locations from two images are the same~\citep{lowryVisualPlaceRecognition2016}.
Together they can be used with Street View Images (SVI), allowing the study of temporal changes from a human perspective \citep{biljeckiStreetViewImagery2021,huangCityPulseFineGrainedAssessment2024}.
\\

However, to quantify temporal changes with SVI, knowing that two images depict the same location is not sufficient, because two images from the same location typically represent the same location but not the identical view~\citep{choEnvironmentalChangeDetection2025}. 
Even if images are matched with GPS data, vehicle routing, such as driving in different lanes or irregular capture intervals can cause visual differences in the photos~\citep{sakuradaChangeDetectionStreet2015}. \\

Deep learning based approaches have emerged as the mainstream way for SCD~\citep{kannanZeroSCDZeroShotStreet2024}.
\cite{sakuradaChangeDetectionStreet2015} used CNNs, superpixel segmentation, and the geometric structure of images to estimate the temporal changes between two images with nearly aligned viewpoints.
\cite{kannanZeroSCDZeroShotStreet2024} developed a zero-shot change detection model with semantic segmentation verification to generate binary change masks. 
Using SVI, \cite{lopesExploringBeforeandAfterVisual2026} explored the before-and-after preferences related to street interventions. 
Both \cite{huangDetectingNeighborhoodGentrification2022} and \cite{xiaoExaminingStateledGentrification2026} studied gentrification across years, while
\cite{naikComputerVisionUncovers2017} studied the perception of safety between 2007 and 2014 from SVI.
\\

\begin{figure}
    \centering
    \includegraphics[width=1\linewidth]{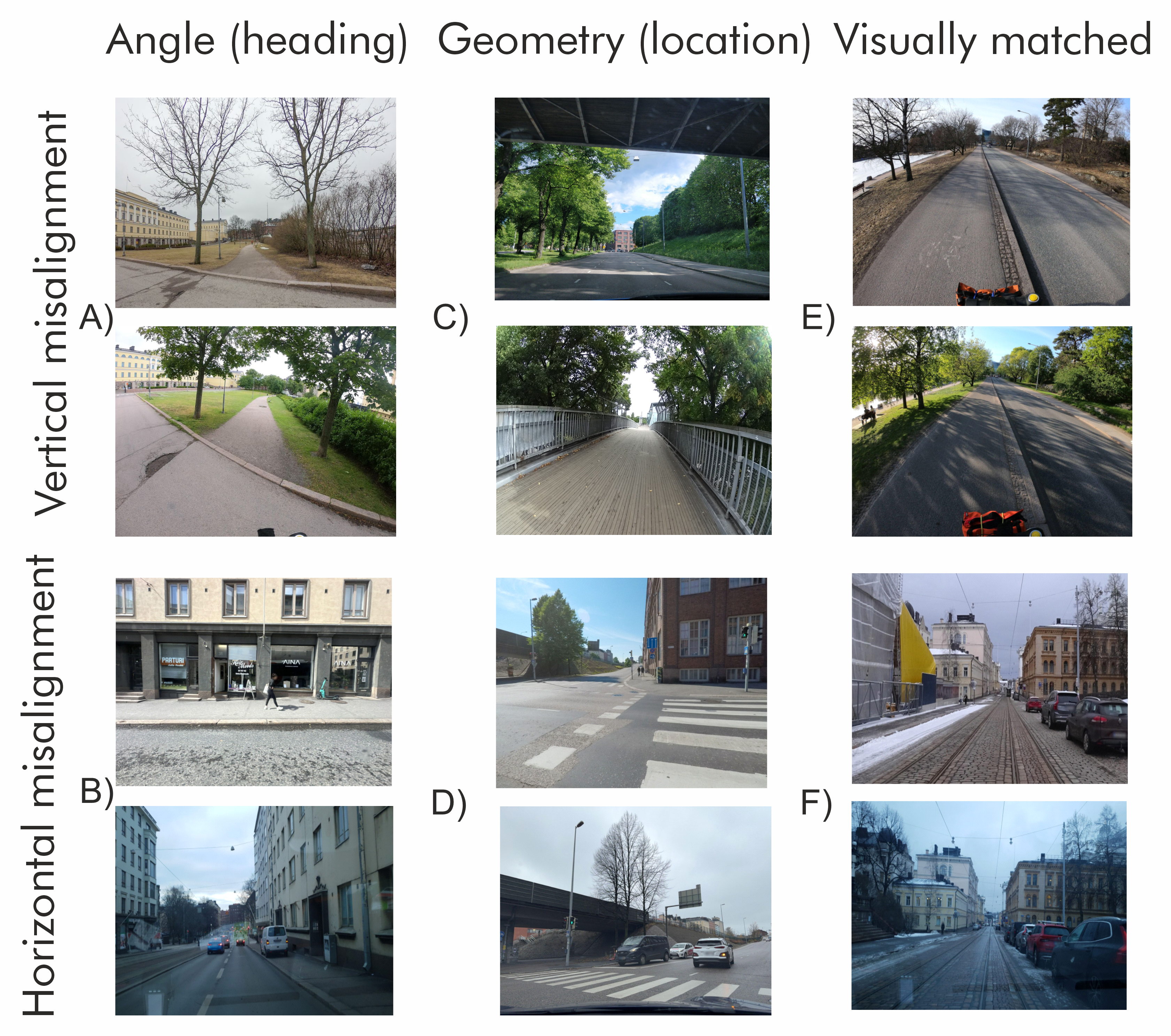}
    \caption{A) Mapillary metadata only considers the horizontal heading angle, leading to one of the images capturing more ground or sky features.
    B) However, the horizontal heading angle is commonly inconsistent, as seen on the example where it is vehicle heading for one image, and the camera heading for the other.
    C) Mapillary geographic coordinates also cannot capture vertical variation, thus one of the images was taken on the bridge above the other image.
    D) Finally, even horizontal geographic coordinates are inaccurate, causing viewpoint changes.
    E) and F) show visually aligned image pairs which have passed strict horizontal and vertical filtering criteria.
    All examples used a 2.5 m geographic distance and a 30-degree maximum difference of the viewing angle from the Mapillary metadata.
    }
    \label{fig:example_of_issues}
\end{figure}

However, in the aforementioned studies, there is a lack of methods for estimating how well aligned the two images from different time periods were.
Instead, they rely only on metadata-based GPS coordinates, ignoring the possible influences from unaligned images.
Current approaches attempt to detect changes of features, but do not consider how well the images match visually with each other.
Scenes can have actual changes and temporary changes (weather, vehicles, pedestrians), but also artifacts derived from the capture itself (positional alignment) \citep{alcantarillaStreetviewChangeDetection2018}.
Thus, if visually unaligned images are compared, deviations may introduce bias and errors for studies trying to evaluate temporal changes between images, as shown in \autoref{fig:example_of_issues}.
They also do not allow estimating whether a pair of images is comparable for a perception study.\\

To tackle these problems, this paper presents the PairWise image finder tool that uses feature detection, feature matching and semantic segmentation to evaluate scene change (SCD) while integrating VPR by evaluating how well two street-level views are aligned.
The approach introduced differs from existing SCD and VPR algorithms, as instead of directly detecting changes or locations, it identifies visually comparable pairs with the same visual alignment.
The tool can help with analyzing temporal changes across months, seasons, and years or find robustly paired images for street-level visual perception studies

\section{Description}

To fill the gap between Scene Change Detection and Visual Place Recognition to understand the visual alignment of two images from nearly the same location, the PairWise image finder delivers a set of user-defined thresholds using computer vision methods.
By varying the thresholds, the user can find visually aligned image pairs suitable for their use case.

\section{Architecture of the tool}

\subsection{Mapillary and SVI metadata} 
By default, the tool runs on Python and can be used after installing the required packages listed in the included requirements.txt.
Firstly, the tool uses the publicly available metadata from Mapillary \citep{mapillaryMapillary2026}, which provides four key variables for starting the matching of each image: \textit{geometry} (a geographic location with longitudinal and latitudinal coordinates), \textit{captured\_at} (the timestamp of capture), \textit{compass\_angle} (the compass angle or heading of the image), and \textit{is\_pano} (whether image is panoramic or not).
These can be used to perform preliminary selection on the data with user-defined limits to reduce the amount of pairs being processed.
The capture time can determine which two temporal periods are compared, the location can limit the pairs assessed to only images near each other, the angle can be used to remove non-panoramic (perspective) images that face the wrong direction from the same location, and the panorama variable can be used to select the type of SVI data suitable for the study at hand.
All of the Mapillary metadata-based filters can be adjusted to allow larger quantity of image pairs enter the subsequent analysis, at the cost of increased processing time.
The tool can also be configured to use other pre-downloaded image sources, such as Google Street View data, if it is provided with a location dataset for the images.

\subsection{Alignment of the images}
However, even with the metadata, there is a need to adjust for the inconsistency in the locations and angles.
For this, the tool uses three main technical solutions: (1) feature detection with SuperPoint~\citep{detoneSuperPointSelfSupervisedInterest2017}, (2) feature matching with LightGlue~\citep{lindenbergerLightGlueLocalFeature2023}, and (3) Semantic Mask comparisons with pretrained OneFormer models \citep{jainOneFormerOneTransformer2022} trained on Cityscapes~\citep{cordtsCityscapesDatasetSemantic2016} or ADE20K~\citep{zhouSceneParsingADE20K2017}. 
Firstly, the SuperPoint algorithm detects key features in the images of the pair, with a default max setting of 2048 features found for each image.
Then, LightGlue matches local features in both images.
With the matched keypoints, the PairWise tool can assess the total proportion of the features matched out of all detected features, and calculate the average distances between matched keypoints, indicating visual alignment.
It will also assess the coverage of the matched keypoints using convex hull with outlier filtering, first by assessing their share of the image area, and afterwards by comparing the Mean Intersection over Union (mIoU).
Afterwards, the semantic model chosen segments both images, allowing mIoU comparisons of the semantic masks between the two images.
More specifically, the tool uses the masks related to street infrastructure, as they provide a good assessment of lane positioning.
However, other semantic masks can also be compared.
All models used at this step are publicly available as pre-trained models from the HuggingFace repository~\citep{huggingfaceinc.HuggingFace2025}, enabling easy usage. 
The amount of matched features, average match distances, and semantic similarity can then be evaluated to determine the visual alignment of the compared images.\\

After the initial assessment of the images, it removes detected outlier features and use the matched keypoints to align the two images and crop both to a shared visual area.
On the other hand, the center of panoramic images often faces the wrong direction and there are large distortions in the bottom and top halves of the images.
The tool can perform horizontal yaw-adjustment on the panoramic images by searching for the yaw with the minimum keypoint distance and crop a user-defined portion of the top and bottom halves of the images (default: 20\% from both top and bottom; \cite{sanchezAccessingEyelevelGreenness2024}).
Based on the semantic maps, the tool can also remove temporary elements such as people and vehicles from both images that might influence the calculations.
If the additional alignment, cropping, and filtering options are on, the tool will calculate the resulting metrics for the adjusted images.
Finally, the tool can calculate basic image descriptor statistics, such as brightness, contrast, sharpness etc., to give additional insights to the similarity of the images. \\
The main features provided by the tool are listed in \autoref{tab:features}.

\begin{table*}
\small\sffamily\centering
\caption{Overview of filtering, alignment, and comparison features.}
\label{tab:features}

\begin{tabular}{p{3cm} p{5cm} p{3.2cm} p{2cm} p{2cm}}
\toprule
Feature & Usage & Filter & Reference & Notes \\
\specialrule{0.10em}{0.3em}{0.3em}

\multicolumn{5}{l}{\textbf{Mapillary metadata}} \\
\midrule
Date & Selecting images from different temporal periods & month, season, year, group of years, year and season & \cite{mapillaryMapillary2026} & \\
Geographic proximity & Preliminary filtering of data & buffered & & \\
Angle of image (perspective only) & Preliminary filtering of data & & & \\
Panoramic & Choosing panoramic, non-panoramic, or any images & & & \\

\specialrule{0.10em}{0.6em}{0.3em}
\multicolumn{5}{l}{\textbf{Computer vision}} \\
\midrule
Feature detection & Find key features & & \cite{detoneSuperPointSelfSupervisedInterest2017} & can ignore sky features with adjustable pixel buffer \\
Feature matching & Match key features across images &
\% of matched features
\newline average pixel diagonal pixel distance
\newline minimum matched feature coverage (convex hull)
\newline mIoU of matched feature coverage
& \cite{lindenbergerLightGlueLocalFeature2023} & Outliers ignored for coverage calculation \\
Semantic mask alignment & Compare semantic segmentation masks & mIoU of road-related classes & \cite{jainOneFormerOneTransformer2022} & \\

\specialrule{0.10em}{0.6em}{0.3em}
\multicolumn{5}{l}{\textbf{Image adjustment}} \\
\midrule
Panorama cropping & Remove distorted areas of panoramas & \% of top or bottom & & \\
Panorama yaw-adjustment & Align panorama centers using matched features & N degree steps & & \\
Perspective alignment & Realign images using matched keypoints & & & \\
Perspective cropping & Crop to matching visual area & & & \\
Temporary cropping & Remove temporary features before semantic comparison & & & Performs identical semantic cropping on both images \\

\specialrule{0.10em}{0.6em}{0.3em}
\multicolumn{5}{l}{\textbf{Image metrics}} \\
\midrule
Brightness, contrast, sharpness, alignment & Additional insights between images & & & \\

\bottomrule
\end{tabular}
\end{table*}

\section{Use case}

To test the tool, it was used on a case study of the longitudinal changes in Helsinki, Finland, between two three-year time periods (2016-2018, 2024-2026) with a five-year gap.
For the test, a random sample of images was first collected from the study area.
Afterwards, the tool was used with three rounds of filtering; first, only using Mapillary metadata (images in a pair within 2.5 m, max 30-degree angle difference), a second one with the metadata and exclusion of the bottom 25\% of pairs based on features matched, feature distance, minimum matched feature coverage, feature coverage mIoU and mIoU of the semantic road classes.
As the final example, the bottom 75\% of pairs were excluded. \\

\begin{figure}
    \centering
    \includegraphics[width=1\linewidth]{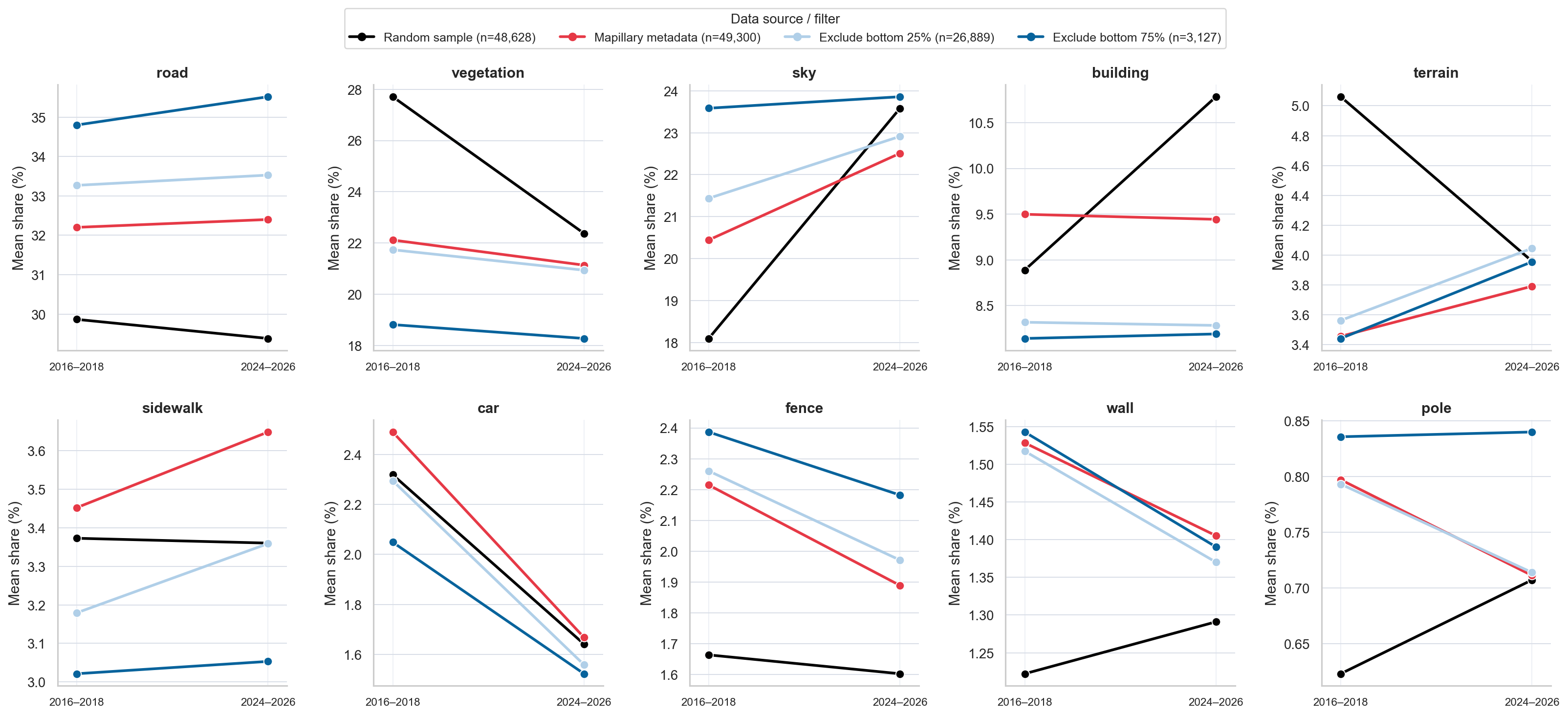}
    \caption{Longitudinal changes in Helsinki, Finland, with a random sample and varying filtering criteria from the tool}
    \label{fig:comparison}
\end{figure}

\autoref{fig:comparison} shows that the patterns of longitudinal change vary between the filtering criteria.
The random sample shows drastic increases for visible sky and buildings, major decreases for visible vegetation and terrain, and minor decreases for visible roads and sidewalks.
However, the filtered pairs show smaller changes in vegetation and sky, while visible terrain increased slightly and the share of visible buildings has stayed the same.
The results highlight the significance of image viewpoints for studying temporal change; the geographic and visual alignment of images influences the patterns discovered.

\section{Discussion}

The tool has certain limitations that need to be considered.
Feature matching has issues with varying illumination conditions~\citep{sakuradaChangeDetectionStreet2015}, making the tool less usable for studying diurnal changes, but also possibly cross-weather comparisons \citep{jiangReviewMultimodalImage2021}.
As the tool heavily relies on the keypoints, drastic changes between two periods of time can limit the amount of possible keypoints and weaken the reliability of the visual alignment assessment, as locations with large changes might be classified as unaligned.
A high volume of SVI and performing looser filtering, combined with additional manual assessments, can help alleviate the issues with misclassifications.
However, the tool has significant computational overhead, with three deep learning models used that can increase the computation time with large datasets.
While stricter preliminary alignment criteria from the Mapillary metadata, which limit the amount of data processed can reduce the processing time, they will also limit the amount of data usable for analysis.
Thus, a balance, depending on the time and resources available is needed for the efficient use of the tool. 
Finally, currently there are no definitive values for well-matched images; the user should select their own thresholds based on the requirements of their use case and available computational power.
\\

The tool could also be improved in the future with additional functionalities.
Currently panoramic and perspective images can be analyzed together or separately, as they offer varying viewpoints and results on the street-level changes~\citep{beaucampFrameEvaluatingPanoramic2025a}. 
However, the tool and the matched keypoints could be used to turn panoramic images into additional perspective images to match with.
The Mapillary metadata could also be further used: for example, \textit{sequence\_id} that contains the sequence of images that one image belongs to, could be used to more efficiently skip matching on pairs which have shown consistent variation from previous images in a sequence.
In the future, more advanced feature matching or AI-based image analysis tools could further improve the alignment process.
As an example, while developing the tool, a SIFT based local feature extractor between paired images~\citep{loweObjectRecognitionLocal1999} was tested, but it did not provide reliable improvements to the currently available methods.

\section{Conclusions}
There has been a gap in understanding how much the visual alignment of street-level images plays a role when assessing temporal changes of locations. 
The PairWise image finder allows users to quantify the visual alignment of nearby image pairs and to select images with required alignment for change and perception studies.

\section*{Data availability}

The full codebase of the PairWise tool and its related document documentation can be found from (\url{https://github.com/jusba/PairWise_image_finder}).

\section*{Acknowledgments}

This work was supported by the Finnish Ministry of Education and Culture’s Pilot for Doctoral Programmes (Pilot project Mathematics of Sensing, Imaging and Modelling) and the Research Council of Finland (Flagship of Advanced Mathematics for Sensing, Imaging and Modelling, FAME, grant number 359182). 
This work was funded by the GREENTRAVEL project by the European Union (ERC, project 101044906). 
Views and opinions expressed are however those of the authors only and do not necessarily reflect those of the European Union or the European Research Council Executive Agency. 
Neither the European Union nor the granting authority can be held responsible for them.
The authors wish to acknowledge CSC – IT Center for Science, Finland, for computational resources of the Puhti supercomputer.

\section*{Declaration of generative AI and AI-assisted technologies in the research process}

During the preparation of this work, the authors used AI assistants for code development and grammar improvement.
After using this tool/service, the authors reviewed and edited the code as needed and accept full responsibility for the content of the published article.

\bibliographystyle{elsarticle-harv}
\bibliography{references}

@article{alcantarillaStreetviewChangeDetection2018,
    title = {Street-view change detection with deconvolutional networks},
    volume = {42},
    issn = {0929-5593, 1573-7527},
    url = {http://link.springer.com/10.1007/s10514-018-9734-5},
    doi = {10.1007/s10514-018-9734-5},
    language = {en},
    number = {7},
    urldate = {2026-04-21},
    journal = {Autonomous Robots},
    author = {Alcantarilla, Pablo F. and Stent, Simon and Ros, Germán and Arroyo, Roberto and Gherardi, Riccardo},
    month = oct,
    year = {2018},
    pages = {1301--1322},
}

@misc{detoneSuperPointSelfSupervisedInterest2017,
    title = {{SuperPoint}: {Self}-{Supervised} {Interest} {Point} {Detection} and {Description}},
    copyright = {arXiv.org perpetual, non-exclusive license},
    shorttitle = {{SuperPoint}},
    url = {https://arxiv.org/abs/1712.07629},
    doi = {10.48550/ARXIV.1712.07629},
    urldate = {2026-05-08},
    publisher = {arXiv},
    author = {DeTone, Daniel and Malisiewicz, Tomasz and Rabinovich, Andrew},
    year = {2017},
    note = {Version Number: 4},
}

@article{lowryVisualPlaceRecognition2016,
    title = {Visual {Place} {Recognition}: {A} {Survey}},
    volume = {32},
    copyright = {https://ieeexplore.ieee.org/Xplorehelp/downloads/license-information/IEEE.html},
    issn = {1552-3098, 1941-0468},
    shorttitle = {Visual {Place} {Recognition}},
    url = {https://ieeexplore.ieee.org/document/7339473/},
    doi = {10.1109/TRO.2015.2496823},
    number = {1},
    urldate = {2026-04-21},
    journal = {IEEE Transactions on Robotics},
    author = {Lowry, Stephanie and Sünderhauf, Niko and Newman, Paul and Leonard, John J. and Cox, David and Corke, Peter and Milford, Michael J.},
    month = feb,
    year = {2016},
    pages = {1--19},
}

@misc{choEnvironmentalChangeDetection2025,
    title = {Environmental {Change} {Detection}: {Toward} a {Practical} {Task} of {Scene} {Change} {Detection}},
    copyright = {arXiv.org perpetual, non-exclusive license},
    shorttitle = {Environmental {Change} {Detection}},
    url = {https://arxiv.org/abs/2506.11481},
    doi = {10.48550/ARXIV.2506.11481},
    urldate = {2026-04-21},
    publisher = {arXiv},
    author = {Cho, Kyusik and Woo, Suhan and Seong, Hongje and Kim, Euntai},
    year = {2025},
    note = {Version Number: 1},
}

@inproceedings{sakuradaChangeDetectionStreet2015,
    address = {Swansea},
    title = {Change {Detection} from a {Street} {Image} {Pair} using {CNN} {Features} and {Superpixel} {Segmentation}},
    isbn = {978-1-901725-53-7},
    url = {http://www.bmva.org/bmvc/2015/papers/paper061/index.html},
    doi = {10.5244/C.29.61},
    language = {en},
    urldate = {2026-05-21},
    booktitle = {Procedings of the {British} {Machine} {Vision} {Conference} 2015},
    publisher = {British Machine Vision Association},
    author = {Sakurada, Ken and Okatani, Takayuki},
    year = {2015},
    pages = {61.1--61.12},
}

@article{naikComputerVisionUncovers2017,
  title = {Computer Vision Uncovers Predictors of Physical Urban Change},
  author = {Naik, Nikhil and Kominers, Scott Duke and Raskar, Ramesh and Glaeser, Edward L. and Hidalgo, C{\'e}sar A.},
  year = 2017,
  month = jul,
  journal = {Proceedings of the National Academy of Sciences},
  volume = {114},
  number = {29},
  pages = {7571--7576},
  issn = {0027-8424, 1091-6490},
  doi = {10.1073/pnas.1619003114},
  urldate = {2026-05-21},
  langid = {english}
}

@inproceedings{loweObjectRecognitionLocal1999,
    address = {Kerkyra, Greece},
    title = {Object recognition from local scale-invariant features},
    isbn = {978-0-7695-0164-2},
    url = {http://ieeexplore.ieee.org/document/790410/},
    doi = {10.1109/ICCV.1999.790410},
    urldate = {2026-05-21},
    booktitle = {Proceedings of the {Seventh} {IEEE} {International} {Conference} on {Computer} {Vision}},
    publisher = {IEEE},
    author = {Lowe, D.G.},
    year = {1999},
    pages = {1150--1157 vol.2},
}

@misc{lindenbergerLightGlueLocalFeature2023,
    title = {{LightGlue}: {Local} {Feature} {Matching} at {Light} {Speed}},
    copyright = {arXiv.org perpetual, non-exclusive license},
    shorttitle = {{LightGlue}},
    url = {https://arxiv.org/abs/2306.13643},
    doi = {10.48550/ARXIV.2306.13643},
    urldate = {2026-05-08},
    publisher = {arXiv},
    author = {Lindenberger, Philipp and Sarlin, Paul-Edouard and Pollefeys, Marc},
    year = {2023},
    note = {Version Number: 1},
}

@inproceedings{cordtsCityscapesDatasetSemantic2016,
    address = {Las Vegas, NV, USA},
    title = {The {Cityscapes} {Dataset} for {Semantic} {Urban} {Scene} {Understanding}},
    isbn = {978-1-4673-8851-1},
    url = {http://ieeexplore.ieee.org/document/7780719/},
    doi = {10.1109/CVPR.2016.350},
    urldate = {2024-05-28},
    booktitle = {2016 {IEEE} {Conference} on {Computer} {Vision} and {Pattern} {Recognition} ({CVPR})},
    publisher = {IEEE},
    author = {Cordts, Marius and Omran, Mohamed and Ramos, Sebastian and Rehfeld, Timo and Enzweiler, Markus and Benenson, Rodrigo and Franke, Uwe and Roth, Stefan and Schiele, Bernt},
    month = jun,
    year = {2016},
    note = {GSCC: 0013884},
    pages = {3213--3223},
}

@inproceedings{zhouSceneParsingADE20K2017,
    address = {Honolulu, HI},
    title = {Scene {Parsing} through {ADE20K} {Dataset}},
    isbn = {978-1-5386-0457-1},
    url = {http://ieeexplore.ieee.org/document/8100027/},
    doi = {10.1109/CVPR.2017.544},
    urldate = {2024-05-28},
    booktitle = {2017 {IEEE} {Conference} on {Computer} {Vision} and {Pattern} {Recognition} ({CVPR})},
    publisher = {IEEE},
    author = {Zhou, Bolei and Zhao, Hang and Puig, Xavier and Fidler, Sanja and Barriuso, Adela and Torralba, Antonio},
    month = jul,
    year = {2017},
    note = {GSCC: 0003146},
    pages = {5122--5130},
}

@misc{mapillaryMapillary2026,
    title = {Mapillary},
    url = {https://www.mapillary.com/},
    language = {en},
    author = {{Mapillary}},
    year = {2026},
    note = {GSCC: 0000491},
}

@misc{kannanZeroSCDZeroShotStreet2024,
    title = {{ZeroSCD}: {Zero}-{Shot} {Street} {Scene} {Change} {Detection}},
    copyright = {Creative Commons Attribution 4.0 International},
    shorttitle = {{ZeroSCD}},
    url = {https://arxiv.org/abs/2409.15255},
    doi = {10.48550/ARXIV.2409.15255},
    urldate = {2026-04-21},
    publisher = {arXiv},
    author = {Kannan, Shyam Sundar and Min, Byung-Cheol},
    year = {2024},
    note = {Version Number: 1},
}

@article{lopesExploringBeforeandAfterVisual2026,
    title = {Exploring {Before}-and-{After} visual preferences of street interventions using {Google} {Street} {View} time travel},
    volume = {273},
    issn = {01692046},
    url = {https://linkinghub.elsevier.com/retrieve/pii/S0169204626001106},
    doi = {10.1016/j.landurbplan.2026.105686},
    language = {en},
    urldate = {2026-05-11},
    journal = {Landscape and Urban Planning},
    author = {Lopes, Maria Inês and Valença, Gabriel and Moura, Filipe},
    month = sep,
    year = {2026},
    pages = {105686},
}

@article{xiaoExaminingStateledGentrification2026,
    title = {Examining state-led gentrification using street view imagery: {Evidence} from {Shanghai}, {China}},
    volume = {175},
    issn = {02642751},
    shorttitle = {Examining state-led gentrification using street view imagery},
    url = {https://linkinghub.elsevier.com/retrieve/pii/S0264275126003458},
    doi = {10.1016/j.cities.2026.107113},
    language = {en},
    urldate = {2026-05-08},
    journal = {Cities},
    author = {Xiao, Yang and Tang, Yiwen and Li, Hong and He, Shenjing},
    month = aug,
    year = {2026},
    pages = {107113},
}

@misc{huangCityPulseFineGrainedAssessment2024,
    title = {{CityPulse}: {Fine}-{Grained} {Assessment} of {Urban} {Change} with {Street} {View} {Time} {Series}},
    copyright = {arXiv.org perpetual, non-exclusive license},
    shorttitle = {{CityPulse}},
    url = {https://arxiv.org/abs/2401.01107},
    doi = {10.48550/ARXIV.2401.01107},
    urldate = {2026-04-21},
    publisher = {arXiv},
    author = {Huang, Tianyuan and Wu, Zejia and Wu, Jiajun and Hwang, Jackelyn and Rajagopal, Ram},
    year = {2024},
    note = {Version Number: 2},
}

@inproceedings{huangDetectingNeighborhoodGentrification2022,
    address = {Osaka, Japan},
    title = {Detecting {Neighborhood} {Gentrification} at {Scale} via {Street}-level {Visual} {Data}},
    copyright = {https://doi.org/10.15223/policy-029},
    isbn = {978-1-6654-8045-1},
    url = {https://ieeexplore.ieee.org/document/10020341/},
    doi = {10.1109/BigData55660.2022.10020341},
    urldate = {2026-05-21},
    booktitle = {2022 {IEEE} {International} {Conference} on {Big} {Data} ({Big} {Data})},
    publisher = {IEEE},
    author = {Huang, Tianyuan and Dai, Timothy and Wang, Zhecheng and Yoon, Hesu and Sheng, Hao and Ng, Andrew Y. and Rajagopal, Ram and Hwang, Jackelyn},
    month = dec,
    year = {2022},
    pages = {1632--1640},
}

@article{biljeckiStreetViewImagery2021,
  title = {Street View Imagery in Urban Analytics and {{GIS}}: {{A}} Review},
  shorttitle = {Street View Imagery in Urban Analytics and {{GIS}}},
  author = {Biljecki, Filip and Ito, Koichi},
  year = 2021,
  month = nov,
  journal = {Landscape and Urban Planning},
  volume = {215},
  pages = {104217},
  issn = {01692046},
  doi = {10.1016/j.landurbplan.2021.104217},
  urldate = {2024-02-28},
  langid = {english}
}

@misc{huggingfaceinc.HuggingFace2025,
  title = {Hugging {{Face}}},
  author = {{Hugging Face, Inc.}},
  year = 2026,
  month = jan,
  journal = {The AI community building the future.},
  urldate = {2025-07-11},
  howpublished = {https://huggingface.co/huggingface}
}

@article{sanchezAccessingEyelevelGreenness2024,
    title = {Accessing {Eye}-level {Greenness} {Visibility} from {Open}-{Source} {Street} {View} {Images}: {A} methodological development and implementation in multi-city and multi-country contexts},
    issn = {22106707},
    shorttitle = {Accessing {Eye}-level {Greenness} {Visibility} from {Open}-{Source} {Street} {View} {Images}},
    url = {https://linkinghub.elsevier.com/retrieve/pii/S221067072400091X},
    doi = {10.1016/j.scs.2024.105262},
    language = {en},
    urldate = {2024-02-12},
    journal = {Sustainable Cities and Society},
    author = {Sánchez, Ilse Abril Vázquez and Labib, Dr. S.M.},
    month = feb,
    year = {2024},
    note = {GSCC: 0000011},
    pages = {105262},
}

@misc{jainOneFormerOneTransformer2022,
    title = {{OneFormer}: {One} {Transformer} to {Rule} {Universal} {Image} {Segmentation}},
    copyright = {arXiv.org perpetual, non-exclusive license},
    shorttitle = {{OneFormer}},
    url = {https://arxiv.org/abs/2211.06220},
    doi = {10.48550/ARXIV.2211.06220},
    urldate = {2025-06-16},
    publisher = {arXiv},
    author = {Jain, Jitesh and Li, Jiachen and Chiu, MangTik and Hassani, Ali and Orlov, Nikita and Shi, Humphrey},
    year = {2022},
    note = {Version Number: 2},
}

@article{beaucampFrameEvaluatingPanoramic2025a,
    title = {Beyond the frame: evaluating panoramic vs. perspective images for assessing place perception},
    volume = {39},
    issn = {1365-8816, 1362-3087},
    shorttitle = {Beyond the frame},
    url = {https://www.tandfonline.com/doi/full/10.1080/13658816.2025.2483857},
    doi = {10.1080/13658816.2025.2483857},
    language = {en},
    number = {10},
    urldate = {2026-05-22},
    journal = {International Journal of Geographical Information Science},
    author = {Beaucamp, Benjamin and Leduc, Thomas and Tourre, Vincent and Servières, Myriam},
    month = oct,
    year = {2025},
    pages = {2300--2332},
}

@article{jiangReviewMultimodalImage2021,
    title = {A review of multimodal image matching: {Methods} and applications},
    volume = {73},
    issn = {15662535},
    shorttitle = {A review of multimodal image matching},
    url = {https://linkinghub.elsevier.com/retrieve/pii/S156625352100035X},
    doi = {10.1016/j.inffus.2021.02.012},
    language = {en},
    urldate = {2026-04-21},
    journal = {Information Fusion},
    author = {Jiang, Xingyu and Ma, Jiayi and Xiao, Guobao and Shao, Zhenfeng and Guo, Xiaojie},
    month = sep,
    year = {2021},
    pages = {22--71},
}

\end{document}